\documentclass{article}

\usepackage[final]{neurips_2023}

\usepackage[utf8]{inputenc} 
\usepackage[T1]{fontenc}    
\usepackage{hyperref}       
\usepackage{url}            
\usepackage{booktabs}       
\usepackage{amsfonts}       
\usepackage{nicefrac}       
\usepackage{microtype}      
\usepackage{xcolor}         
\usepackage{booktabs}
\usepackage{booktabs}
\usepackage{multirow}
\usepackage{algorithmic}
\usepackage{graphicx}
\usepackage{amsmath}
\usepackage{algorithm}

\title{Leading the Pack: N-player Opponent Shaping}
\newcommand{\ucl}{1}
\newcommand{\oxford}{2}

\author{%
    \textbf{Alexandra Souly}$^{ \ucl{}}$
    \textbf{Timon Willi}$^\ucl{}$
    \textbf{Akbir Khan}$^\oxford{}$
    \textbf{Robert Kirk}$^\ucl{}$\\
    \textbf{Chris Lu}$^\oxford{}$
    \textbf{Edward Grefenstette}$^\ucl{}$
    \textbf{Tim Rockt{\"a}schel}$^\ucl{}$\\[0.25em]
    ${}^\ucl{}$University College London ~${}^\oxford{}$University of Oxford\\[0.25em]
    \texttt{alexandrasouly@gmail.com}
}

\begin{document}

\maketitle

\begin{abstract}
Reinforcement learning solutions have great success in the 2-player general sum setting. In this setting, the paradigm of Opponent Shaping (OS), in which agents account for the learning of their co-players, has led to agents which are able to avoid collectively bad outcomes, whilst also maximizing their reward. These methods have currently been limited to 2-player game. However, the real world involves interactions with many more agents, with interactions on both local and global scales. In this paper, we extend Opponent Shaping (OS) methods to environments involving multiple co-players and multiple shaping agents. We evaluate on over 4 different environments, varying the number of players from 3 to 5, and demonstrate that model-based OS methods converge to equilibrium with better global welfare than naive learning. However, we find that when playing with a large number of co-players, OS methods' relative performance reduces, suggesting that in the limit OS methods may not perform well. Finally, we explore scenarios where more than one OS method is present, noticing that within games requiring a majority of cooperating agents, OS methods converge to outcomes with poor global welfare.
\end{abstract}

\section{Introduction}

Multi agent systems are increasingly deployed to the real world, in which interactions may be either cooperative, competitive or both. These can be considered general-sum games where interactions combine opportunities for coordination with self-interested motivations, where each player is trying to maximise their own reward. Real-world examples of such games include global pollution, deforestation, over-fishing and arms races. In these settings, agents are incentivised to exploit the system and/or other players for their own gains by defecting, but cooperation would result in greater social welfare.

General sum games have multiple equilibria, many of which do not produce high social welfare. Naive learning algorithms that work well in zero-sum settings do not perform well in general-sum settings: they learn the Nash-equilibrium strategies by best-responding to the opponent's past behaviour, but in these settings many Nash-equilibria often coincides with worst-case outcomes for all participants \citep{sos}. 

Opponent shaping (OS) methods are a family of approaches that take into account the evolving social dynamics instead of simply best-responding to the opponent's behaviour \citep{lola, lu2022model, pola, cola, khan2023context}. They act to influence other participant's learning dynamics to result in a favorable equilibrium. These methods have been successfully used in 2-player general sum settings to achieve high rewards, however they do not always achieve pro-social outcomes, and can sometimes successfully exploit their opponent.

The study of OS methods has so far been limited to only two interacting agents, whilst the real world has many more agents. The interactions of more players give rise to more complicated and unstable social dynamics, making cooperation harder to achieve. In particular, in diverse populations equilibrium selection requires modelling not only a single co-player's beliefs but also their beliefs of other co-players.

\begin{figure}
    \centering
    \includegraphics[width=\textwidth]{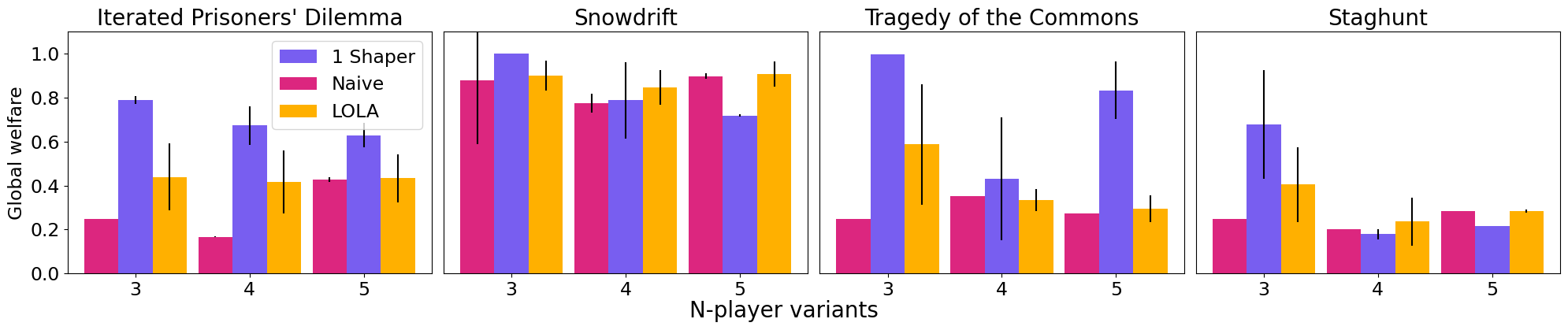}
    \caption{\textbf{Normalised
    global welfare with a single shaping agent.} We report the converged (normalised) global welfare for training a shaping agent with unseen co-players across games and player variants. Across most games \textsc{shaper} outperforms LOLA and as number of players increase, both OS methods fail to find better welfare maximising equilibria than the naive baseline. Results are reported over 5 seeds with mean and standard deviation.}
    \label{fig:single-shaping}
\end{figure}
In this work, we introduce a natural extension for OS methods to more-than-2-player setting. We then evaluate this extension over a total of 12 varying games. We find that OS methods consistently help achieve higher global welfare than baseline methods. While OS-methods are only maximising their own reward, they do so while effectively avoiding collective welfare minimising outcomes, creating better social dynamics and thus also improving their co-players' rewards. We also find that between model-based and model-free OS methods, model-free methods perform better in games with more players. In some instances where OS methods are unable to exploit their opponents, they are also willing to take lower rewards in order to avoid a collectively bad outcome. This is promising for scenarios that involve some more capable self-interested agents in a population of less capable agents.
\section{Background}

Model-based OS methods such as LOLA \cite{lola}, COLA \cite{cola}, POLA \cite{pola}, SOS \cite{sos} and Meta-MAPG \cite{kim2021policy} assume access to the other agents parameters, and as such, anticipate their co-player's updates as they learn. For example, LOLA conditions its update on a co-player's next set of parameters (effectively thinking one step ahead). COLA, POLA and SOS are further adaptations and extensions of LOLA, addressing different shortcomings of the original method. Meta-MAPG also differentiates through the opponent's current parameters via an additional peer learning objective. 

In comparison, model-free OS methods \citep{lu2022model, khan2023context, fung2023analyzing,lu2023adversarial} frame opponent-shaping as a meta-reinforcement learning problem, where the goal is to maximise reward over a series of games against a co-player that is learning. In these approaches a shaping agent trains over $T$ \textit{trials} consisting of $E$ \textit{episodes}. Each trial consists of the following: the shaping agent plays the game with the co-player for $E$ episodes, with the co-player updating after every episode according to its update rule. After every episode, the shaping agent updates its hidden state to add the played games into its memory, whilst the opponent takes a full gradient learning step. After a trial the shaping agent updates its policy parameters. 
Importantly, the shaping agent does not update during an episode unlike the co-player.
\section{Methods}

We expand \textsc{shaper} \citet{khan2023context} to a multi-player setting and generalise the algorithm to arbitrary numbers of shaping-agents and co-players. Co-players can be any type of other agents, including \textsc{shaper}, other meta-agents or naive learners.

We extend opponent-shaping to a generic number of shaping agents and co-players naturally. Similar to single \textsc{shaper} single co-player shaping, shaping agents play the game with their co-players for $E$ episodes but with all co-players simultaneously updating after an episode. After every episode, all shaping agents simultaneously update their respective hidden states. Shaping agents learn to collectively shape their naive co-players.

We update \textsc{shaper}s using evolutionary strategies \citep{salimans2017evolution}, which are computationally better suited to the long timescale of trials and are efficiently parallelisable. Given a game with $N$ \textsc{shaper}s and $K$ co-players, at the start of a trial we sample a population of groups of \textsc{shaper}s. These groups then each play against copies of the same co-players. At the end of the trial, we sample a new generation of groups based on the parameters of the most successful groups. Importantly, there is no mixture between the groups. \textsc{shaper}s do not play with members of other groups, only with the \textsc{shaper}s in their own group, therefore the whole group evolves together. We present a multiplayer extension in Appendix \ref{app}. 

\section{Experiments}

For all games, one episode consists of 100 iterated cooperate-defect choices. One trial consists of 1000 episodes.
We evaluate on the following general sum games:

\textbf{Iterated Prisoner's Dilemma (IPD)}
is a game in which agents must commit to cooperating (staying silent) or defecting (implicating your comrades) \citep{nipd:yao}. For each player, assuming that the actions of others players is fixed, defecting always results in a better payoff than cooperating (that is to say, defection is the dominant choice). All players choosing cooperation is global welfare maximising, but unstable with naive learners. 

\textbf{Snowdrift} is a game in which players choose to shovel (cooperate) or ignore the snow (defect) \citep{snowdrift-nplayer}. The shovelling players share a fixed cost. If a player doesn't shovel, they don't incur a cost but still receive the benefits of the snow is shovelled. If no one shovels, all players receive a low payoff.
This game differs from IPD in that mutual defection is not the best choice - if no one else cooperates, it is still better for the player to shovel alone than if everyone defects and the road stays blocked.

\textbf{Tragedy of the Commons (ToC)} is an N-player social dilemma. We use a general formulation in which there is a social benefit B such that if enough players pay a cost C, everyone gets the benefit. The social benefit can be thought of as the shared resource, and the cost as the opportunity loss of not exploiting the resource. This game encourages free-riding, as long as enough of the co-players cooperate to prevent the tragedy.

\textbf{Stag Hunt} N players choose either to hunt a stag (cooperate) or to hunt a hare (defect) \citep{SH:nplayer}. A certain number of hunters are required to successfully hunt a stag (in our case more than half the total population), the hunt is unsuccessful otherwise. If there is a successful hunt, the payoff is related to the number of cooperating hunters (more hunters implies bigger game). The stag hunters pay the cost of the hunt but everyone equally shares the benefits if the hunt is successful. This setup also encourages free-riding similarly to ToC.
 
\begin{figure}
    \centering
\includegraphics[width=\textwidth]{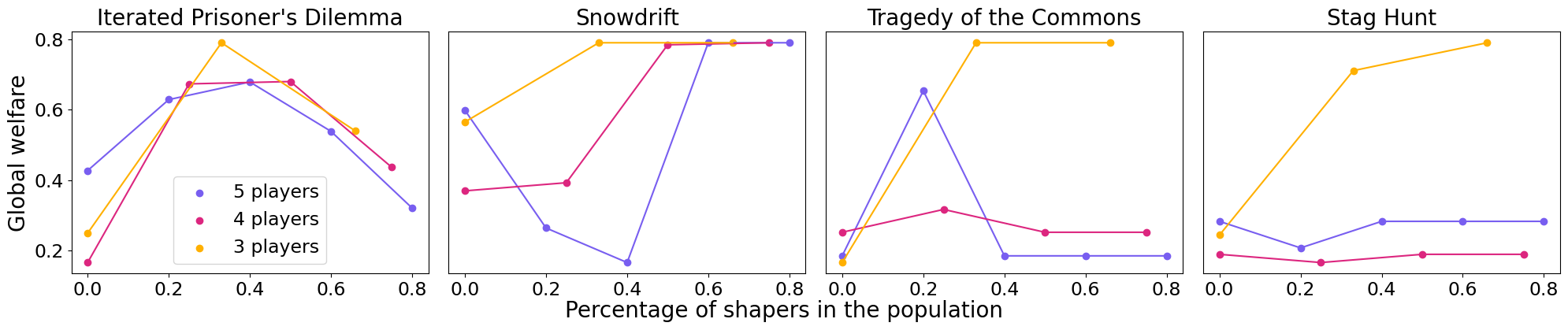}    \caption{\textbf{Normalised global welfare with multiple shaping agents.} In IPD, the optimal amount of \textsc{shaper}s is around a third of the population, too many decreases global welfare. In comparison in Snowdrift - a game where only a single agent is required to cooperate, more shaping agents help coordinate which naive learner is exploited. In ToC and Stag Hunt, where cooperation requires a majority of players, increasing the number of \textsc{shaper}s did not increase welfare. }
    \label{fig:mutli-shaper}
\end{figure}

\begin{figure}
    \centering
\includegraphics[width=\textwidth]{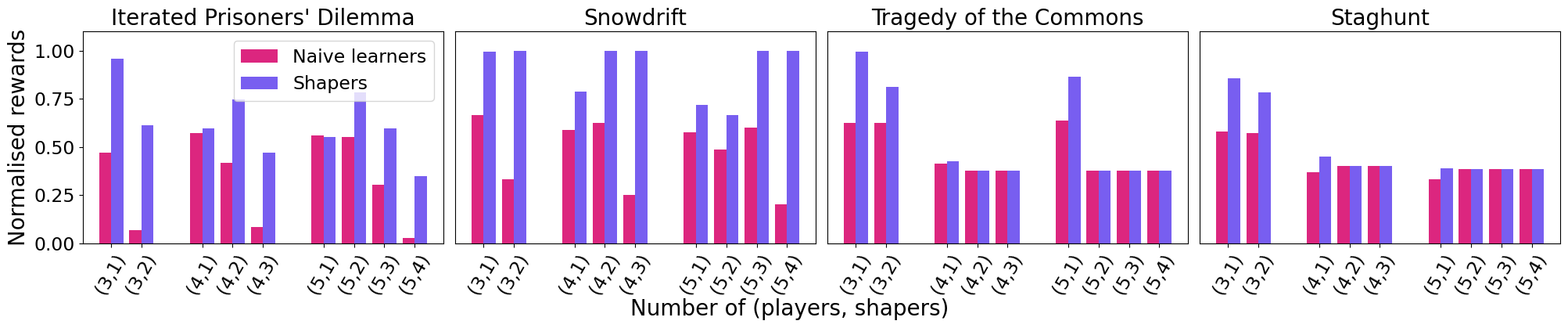}    \caption{\textbf{Normalised \textsc{shaper} versus naive co-player welfare.} We report the (normalised) converged scores attained by \textsc{shaper}s and their naive co-players. In all games, \textsc{shaper}s either score similarly to their naive co-players, or heavily exploit them, particularly in IPD and Snowdrift.}
    \label{fig:exploitation}
\end{figure}

\textbf{Methods} - We consider three different methods for shaping: naive learners, a model-based OS method (LOLA) and the model-free OS method \textsc{shaper}. Naive learners (NLs) are agents that don't account for the learning or adaptation of other agents in the environment. Updates are myopic, at the end of single episodes, and make no modelling assumptions of other players. Our naive learners are parameterized by a recurrent neural network and keep their hidden states between episodes. Their networks are updated using proximal policy optimisation (PPO) \cite{schulman2017proximal} and trained until convergence. For our model-free OS method we use LOLA-PG - a policy gradient approximation to estimate the co-player's gradient update. We use the Infinitely Differentiable Monte Carlo Estimator \citep[DICE]{DICE} to approximate the gradients, and train until convergence. For our model free OS method we use \textsc{shaper}, a meta-agent parameterised by a LSTM network, specifically a GRU with hidden state of size 16. Models are trained until convergence.
\section{Results, Discussion \& Conclusion}
\textbf{Single Shaping Agent} We firstly conduct experiments in which we evaluate a single \textsc{shaper} with $N-1$ naive learners across the many environments. In line with prior work \citep{lu2022model} we find that \textsc{shaper} outperforms LOLA across most tasks in improving global welfare (see Figure \ref{fig:single-shaping}). We also find that current OS methods provide diminishing results when applied on games with more than 2 players. In all 3-player settings, \textsc{shaper} exploits its co-players, but has varying success with more co-players (Figure \ref{fig:exploitation}). In particular, \textsc{shaper} has difficulties on Staghunt and ToC, in scenarios where a majority of players is required for cooperation to be rewarded. 

We note that in ToC the difficulty of cooperation is not monotonic in number of players. In the 4-player game, 3 are required to coordinate to get welfare maximising outcome, whilst in the 5-player variant the required number is also 3. Hence, with randomly searching agents it is easier in the 5-player variant to discover welfare maximising outcomes than the 4-player one. (50\% vs 31\%).

\textbf{Many Shaping Agents} With many shaping agents, we explore the stability of OS methods when interacting with one another. In games where cooperation is related to only a small percentage of a population cooperating, more \textsc{shaper}s improves global welfare (see Figure \ref{fig:mutli-shaper}). In Snowdrift, only a single agent must cooperate, therefore increasing the number of \textsc{shaper}s increases performance as exploiting the naive learners becomes easier. Conversely, in IPD, where incentives for cooperation are lower, we find multiple \textsc{shaper}s reduce global welfare when the majority of the population become \textsc{shaper}s, due to overly exploiting the remaining naive learners (Figure \ref{fig:exploitation}). In games such as ToC and Staghunt, cooperation is only beneficial if enough players coordinate. Two \textsc{shaper}s successfully coordinate in the 3 player game (where only one player can defect), with one \textsc{shaper} cooperating and the other defecting. However they do not succeed at coordination when more players or \textsc{shaper}s are present, suggesting a limitation to the paradigm.

In conclusion, while OS methods improve global welfare over naive learning in multi-agent games, they might do so by exploitation of their co-players. When the game dynamics do not allow for this, they are willing to take lower rewards by cooperating. This is promising for improving the dynamics of finite-resource games like ToC by including more capable agents in the population. The method, however, faces limitations as the number of agents and \textsc{shaper}s increases, especially in games requiring majority cooperation. Future work could extend this to the STORM environments \citep{pax, rutherford2023jaxmarl} with partial observability and temporally-extended action spaces.
\bibliography{refs}

\appendix
\newpage
\section{Additional Environment Details} 

For each game, we present payoff matrices for their work.

\textbf{IPD}
Global welfare increases linearly with the number of cooperators, but agents are incentivised to defect for individually higher payoffs. Therefore this game presents a free-riding problem.
The payoff matrix we use is structured as in Table \ref{tab:payoff_ipd}: 
defection results in 1 more point than cooperation given the actions of the co-players, and for each co-player cooperating our reward increases by 2.
\begin{table}[H]
\centering
\begin{tabular}{lllllll}
\multicolumn{7}{l}{\textbf{No. of C among remaining n-1 players}} \\
     & \multicolumn{1}{c}{}     &$ $0     & $1$     & $2$     & ...    & $n-1$    \\ \cline{3-7} 
\multirow{2}{*}{\textbf{Player A}} &
  \multicolumn{1}{l|}{\textbf{C}} &
  \multicolumn{1}{l|}{$0$} &
  \multicolumn{1}{l|}{$2$} &
  \multicolumn{1}{l|}{$4$} &
  \multicolumn{1}{l|}{...} &
  \multicolumn{1}{l|}{$2(n-1)$} \\ \cline{3-7} 
 &
  \multicolumn{1}{l|}{\textbf{D}} &
  \multicolumn{1}{l|}{$1$} &
  \multicolumn{1}{l|}{$3$} &
  \multicolumn{1}{l|}{$5$} &
  \multicolumn{1}{l|}{...} &
  \multicolumn{1}{l|}{$2n-1$} \\ \cline{3-7} 
\end{tabular}
\caption{N-player IPD payoff where n is the total number of players, C is cooperation action and D is defection action.}
\label{tab:payoff_ipd}
\end{table}

\textbf{StagHunt}

The payoffs are calculated with the following equations, where N is the population size, and the  reward is more than the individual cost of the hunt:

\begin{align*}
\text{if} \quad \text{num coop} &\geq \lceil\frac{N}{2}\rceil:\\
\text{C payoff} &=  \frac{\text{num coop} *  \text{reward}}{N} - \text{hunt cost} \\
\text{D payoff} &=   \frac{\text{num coop} *  \text{reward}}{N} \\
\text{otherwise:} \quad &\\
\text{C payoff} &=  - \text{hunt cost} \\
\text{D payoff} &=  0 \\
\end{align*}
In our experiments we set the cost of the hunt to 3 and the reward for a successful hunt to 6.

\textbf{Tragedy of the Commons}

The payoff matrix we use is presented in Table \ref{tab:payoff_toc}, as per \cite{sep-prisoner-dilemma}. In our experiments, we set the benefit to 5 and the cost to 3.

\begin{table}[H]
\centering
\begin{tabular}{l|c|c|}
\cline{2-3}
                                 & \multicolumn{1}{l|}{\textbf{more than T choose C}} & \multicolumn{1}{l|}{\textbf{T or fewer choose C}} \\ \hline
\multicolumn{1}{|l|}{\textbf{C}} & benefit - cost                                              & -cost                                                \\ \hline
\multicolumn{1}{|l|}{\textbf{D}} & benefit                                                 & 0                                                 \\ \hline
\end{tabular}
\caption{Tragedy of the Commons payoff where T is the threshold for resource exhaustion, C is cooperation action and D is defection action}
\label{tab:payoff_toc}
\end{table}

\section{Multi-Shaping Method}
\label{app}

Following algorithm:
\begin{algorithm}[H] 
\caption{Training multiple \textsc{shaper} agents $\{S\}$ against co-players $\{-S\}$ }\label{alg:cap}. 
\begin{algorithmic}[1]
\REQUIRE Stochastic game $\mathcal{M}$
\STATE Initialise all \textsc{shaper} parameters $\phi_{i}$ for $i \in \{S\}$
\FOR{$t=0$ \textbf{to} $T$}
    \STATE Initialise trial reward $\bar{J}=0$
    \STATE Initialise co-players $\phi_{i}$ for $i \in \{-S\}$
    \STATE{Initialise all agent hidden state $h_i=\textbf{0}$ for $i \in \{S\}\bigcup \{-S\}$}
    
    \FOR{$e=0$ \textbf{to} $E$}
    \STATE Roll out trajectories  $J_{i},  h'_{i} $ = $\mathcal{M}(\phi_{\{S\}}, \phi_{-\{S\}}, h_{\{S\}}, h_{-\{S\}})$ with $i \in \{S\}\bigcup \{-S\}$
    \STATE Update co-player's parameters $\phi_{-i}$ with $i \in \{-S\}$
    \STATE Update all hidden states $h_{i} \leftarrow{} h'_{i}$  with $i \in \{S\}\bigcup \{-S\}$
    \STATE Update trial reward $\bar{J} \leftarrow \bar{J} + J_{i}$ 
    \ENDFOR{} \\
    \STATE{Update \textsc{shaper} parameters $\phi_{i}$ with respect to $\bar{J}$ with $i \in \{S\}$}
    \ENDFOR{} 
\end{algorithmic}
\end{algorithm}

\section{Hyperparameters}
\begin{table}[h]
\parbox[b]{.55\linewidth}{
\centering
\begin{tabular}{l|l}
Hyperparameter & Value \\ \hline
Number of Actor Hidden Layers & 1 \\
Number of Critic Hidden Layers & 1 \\
Torso GRU Size & [16] \\
Length of Trial  & 1000 \\
Length of Episode & 100 \\
Number of Generations & until convergence\\
Number of parallel environments & 2 \\
Number of parallel opponents & 10 \\
Population Size  & 100\\
OpenES sigma init& 0.04    \\
OpenES sigma decay & 0.999 \\
OpenES sigma limit& 0.01  \\
OpenES init min& 0.0  \\
OpenES init max& 0.0 \\ 
OpenES clip min& -1e10  \\  
OpenES clip max& 1e10     \\ 
OpenES lrate init& 0.01    \\
OpenES lrate decay& 0.9999 \\
OpenES lrate limit& 0.001 \\
OpenES beta 1& 0.99       \\
OpenES beta 2& 0.999       \\
OpenES eps& 1e-8          \\
\end{tabular}
\caption{Hyperparameters for SHAPER}}
\parbox[b]{.55\linewidth}{
\centering
\begin{tabular}{l|l}
Hyperparameter & Value \\ \hline
Number of Minibatches & 10\\
Number of Epochs  & 4\\
Gamma & 0.96\\
GAE Lambda & 0.95\\
PPO clipping epsilon& 0.2\\
Value Coefficient& 0.5\\
Clip Value& True \\
Max Gradient Norm& 0.5\\
Anneal Entropy & False\\
Entropy Coefficient Start& 0.1\\
Entropy Coefficient Horizon& 400000000\\
Entropy Coefficient End& 0.01\\
LR Scheduling& False\\
Learning Rate& 0.0003\\
ADAM Epsilon& 1e-5\\
With CNN& False\\
\end{tabular}
\caption{Hyperparameters for PPO}
}
\end{table}

\end{document}